# 3D LiDAR Aided GNSS NLOS Mitigation in Urban Canyons

Weisong Wen, and Li-Ta Hsu*

*Abstract*— **In this paper, we propose a 3D LiDAR aided global navigation satellite system (GNSS) non-line-of-sight (NLOS) mitigation method caused by both static buildings and dynamic objects. A sliding window map describing the surrounding of the ego-vehicle is first generated, based on real-time 3D point clouds from a 3D LiDAR sensor. Then, NLOS receptions are detected based on the sliding window map using a proposed fast searching method which is free of the initial guess of the position of the GNSS receiver. Instead of directly excluding the detected NLOS satellites from further positioning estimation, this paper rectifies the pseudorange measurement model by (1) correcting the pseudorange measurements if the reflecting point of NLOS signals is detected inside the sliding window map, and (2) remodeling the uncertainty of the NLOS pseudorange measurement using a novel weighting scheme. We evaluated the performance of the proposed method in several typical urban canyons in Hong Kong using an automobile-level GNSS receiver. Moreover, we also evaluate the potential of the proposed NLOS mitigation method in GNSS and inertial navigation systems integration via factor graph optimization.**

## I. INTRODUCTION

Positioning in urban environments [1, 2] is becoming essential due to the increasing demand for autonomous driving vehicles (ADV) [3]. The global navigation satellite system (GNSS) [4] is currently one of the principal means of providing globally-referenced positioning for ADV localization. With the increased availability of multiple satellite constellations, GNSS can provide satisfactory performance in open-sky areas [1]. However, the positioning accuracy is significantly degraded in highly-urbanized cities such as Hong Kong, due to signal reflection caused by static buildings [5] and dynamic objects [6] such as double-decker buses. If the direct light-of-sight (LOS) is blocked, and reflected signals from the same satellite are received, the notorious non-light-of-sight (NLOS) receptions occur. According to a recent review paper [7], NLOS is currently the major difficulty in the use of GNSS in intelligent transportation systems. Because of NLOS receptions, the performance of GNSS positioning is highly influenced by real-time surrounding environmental features, such as buildings and dynamic objects. Therefore, effectively sensing and understanding surrounding environments is the key to improving GNSS positioning in urban areas, as GNSS positioning relies heavily on sky view visibility. The most well-known method to cope with the GNSS NLOS receptions is the 3D mapping aided (3DMA) GNSS positioning, such as the NLOS exclusion based on 3D mapping information [8-10], shadow matching [11-13]. However, these 3DMA GNSS methods had the drawbacks of 1) reliance on the availability of 3D building models and the initial guess of the GNSS receiver's position; 2) the inability to mitigate NLOS receptions caused by surrounding dynamic objects. The recent progress of the 3DMA GNSS positioning methods was detailly reviewed in our previous work [14].

Recently, 3D LiDAR sensors, the so-called "eyes" of ADV, the typical indispensable onboard sensor for autonomous driving vehicles, had been employed to detect NLOS caused by dynamic objects in our previous work [6]. Due to the limited field of view (FOV: $-30°{\sim}+10°$ for Velodyne HDL 32E [15]) of 3D LiDAR, only part of a double-decker bus can be scanned. Moreover, the method [6] relies heavily on the accuracy of object detection. However, to the best of our knowledge, this was the first work that employed real-time object detection to help GNSS positioning. Instead of only detecting dynamic objects, the detection of surrounding buildings using real-time 3D LiDAR point clouds [16] was explored. Due to the limited field of view of 3D LiDAR, only part of the buildings can be scanned. Therefore, information about building height was required to detect the NLOS receptions caused by the buildings. Instead of excluding the detected NLOS receptions, we explored ways in which to correct NLOS pseudorange measurements with the help of LiDAR. The 3D LiDAR can measure the distance from the GNSS receiver to the surface of a building that may have reflected the GNSS signal. Then the corrected and the remaining healthy GNSS measurements can both be used in further GNSS positioning. The improved performance was obtained after correcting the detected NLOS satellites [16]. Unfortunately, the performance of this approach relies on the accuracy of the detection of buildings and reflectors. Both building detection and reflector detection can fail when a building surface is highly irregular. The limited FOV (field of view) of LiDAR remains a drawback in the detection of both dynamic objects and buildings. Overall, the work reported in [6] and [16] shows the feasibility of detecting GNSS NLOS using real-time onboard sensing: the real-time point clouds. To overcome the drawback of the limited FOV of 3D LiDAR, we explored the use of both fish-eye cameras and 3D LiDAR to detect and correct NLOS signals [17]. The fish-eye camera was applied to detect NLOS signals. Meanwhile, the 3D LiDAR was employed to measure the

Weisong Wen and L.T. Hsu, are with Hong Kong Polytechnic University, Hong Kong (correspondence e-mail: lt.hsu@polyu.edu.hk).

distance between the GNSS receiver and a potential reflector causing NLOS signals. However, this approach shared similar problems with the work described in [18, 19] where the NLOS detection was sensitive to the environmental illumination conditions. In short, 3D LiDAR aided GNSS positioning is a promising solution for mitigating the effects of NLOS receptions and has several advantages: (1) both dynamic and static objects can be considered during NLOS detection; (2) NLOS detection does not rely on an initial guess of the position of the GNSS receiver; (3) the approach does not require the use of 3D building models, and 3D LiDAR is robust against illumination conditions. However, there are still three major drawbacks: (1) **Small FOV**: the limited FOV of 3D LiDAR causes limited environment sensing capability; (2) **Reliance on the object detection**: the performance of NLOS detection relies heavily on the accuracy of object detection, such as double-decker bus detection and building detection; and (3) **Insufficiency in reflector detection**: the reflector detection method described in [16] can only work when a building surface is detected and is sufficiently regular.

The recently developed LiDAR-based odometry [20, 21] enabled the registration of 3D point clouds from multiple frames into a map. As a result, the FOV of the environment reconstruction can be significantly enhanced by accumulating 3D point clouds from multiple frames. Inspired by this, this paper aims to alleviate the three listed drawbacks of the previously investigated 3D LiDAR aided GNSS positioning methods [6, 16, 17] by exploiting the potential of 3D point clouds from multiple frames. The main contributions of this paper are listed as follows:

(1) **Increase the FOV of 3D LiDAR for NLOS mitigation**: An innovative sliding window map (SWM) was generated based on real-time 3D point clouds from 3D LiDAR. Only the 3D point clouds inside a sliding window are employed to generate the SWM, as the point clouds far away from the GNSS receiver are not necessary for NLOS detection. Therefore, the magnitude of the drift of the SWM is bound to a small value. Note the SWM is generated in real-time for GNSS NLOS detection. As a result, the environment description capability of SWM is significantly better than that of 3D real-time point clouds. Therefore, the FOV of LiDAR sensing is effectively enhanced (*alleviate drawback 1*) which is our first contribution.

(2) **Direct GNSS NLOS identification free of object detection**: As the generated SWM is in the body frame, which is located at the center of the 3D LiDAR, the orientation is directly adopted from an attitude and heading reference system (AHRS) to transform the SWM from the body frame to a local (ENU) frame [22]. Different from the previous work in [6] and [16] which required the object detection algorithm to recover the actual height of the detected dynamic object or the building surface, both the FOV and the density of the environment reconstruction using SWM is significantly enhanced in this paper. Therefore, the NLOS receptions are directly detected based on real-time SWM, which does not require the object detection process, using a fast searching method (*relax the drawback 2*). More importantly, the proposed NLOS detection method does not rely on the initial guess of the GNSS receiver.

(3) **Direct GNSS NLOS reflector detection**: Instead of directly excluding detected NLOS satellites from use in further positioning estimation, this work proposes an approach to rectify the pseudorange measurement model by (1) correcting the pseudorange measurements if the reflecting point of the NLOS signals is detected based on a constrained searching method (*relax the drawback 3*) inside the dense SWM; and (2) re-modeling the uncertainty of NLOS pseudorange measurement using a novel weighting scheme if the reflector is not detected. Different from the previous work in [16] where the reflector detection algorithm requires the almost parallel distribution and the regular surface of the buildings, this paper directly searches the potential reflectors based on the geometry captured by the SWM. Finally, both the corrected and healthy pseudorange measurements are used in GNSS single-point positioning (SPP).

(4) **Experimental verification**: This paper verifies the effectiveness of the proposed method with several challenging datasets collected in urban canyons of Hong Kong. Moreover, the potential of the proposed method in the integration of GNSS and inertial navigation system (INS) is also evaluated.

The remainder of this paper is organized as follows. An overview of the proposed method is given in Section II. The generation of the sliding window map is elaborated in Section III. In Section IV, the proposed NLOS detection, NLOS correction, and remodeling approaches are presented. Several real experiments were performed to evaluate the effectiveness of the proposed method in Section V. Finally, conclusions are drawn, and future work is presented in Section VI.

## II. SYSTEM OVERVIEW AND NOTATIONS

An overview of the proposed method in this paper is shown in Fig. 1. The system consists of two parts: (1) the real-time SWM generation based on clouds from 3D LiDAR and an AHRS, corresponding to the light blue shaded boxes in Fig. 1. (2) the GNSS NLOS detection and rectification (corresponding to the light orange shaded boxes) based on the real-time environment description (the SWM). In this paper, matrices are denoted as uppercase with bold letters, e.g. $\mathbf{G}$. Vectors are denoted as lowercase with bold letters, e.g. $\mathbf{v}$. Variable scalars are denoted as lowercase italic letters, e.g. $t$. Constant scalars are denoted as lowercase letters, e.g. n. Meanwhile, the state of the GNSS receiver and the position of satellites are all expressed in the earth-centered, earth-fixed (ECEF) frame.

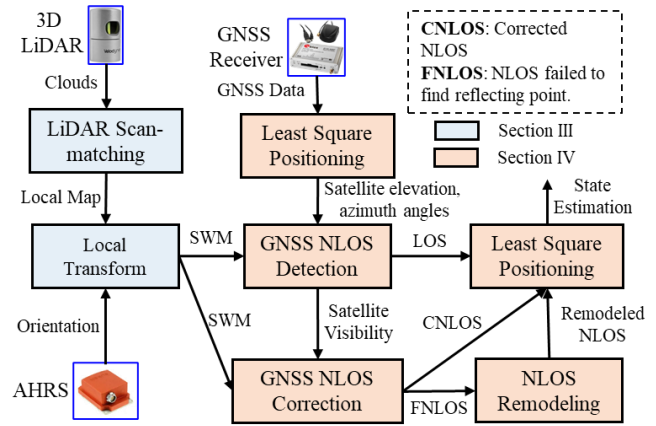

Fig. 1. Overview of the proposed method. The CNLOS denotes the detected and corrected GNSS NLOS measurements. The FNLOS denotes the detected but the reflecting point is not found in the SWM.

 

Be noted that 3D LiDAR is only used for describing/reconstructing the environment and the LiDAR odometry is not integrated with GNSS yet which is similar to our previous work in [16, 17]. To make the proposed pipeline clear, the major notations are defined in Table 1 and followed by the rest of the paper. The variable expressed in the ECEF or east, north, and up (ENU) frames is denoted by superscripts "$G$", "$L$". For example, the transformation from the ENU and the ECEF frame is defined as $\mathbf{T}_L^G = [\mathbf{R}_L^G \quad \mathbf{t}_L^G]$, where the $\mathbf{R}_L^G$ and the $\mathbf{t}_L^G$ denote the rotation and translation, respectively. The body frames of AHRS, LiDAR, and GNSS receiver are denoted by superscripts "$BI$", "$BL$", and "$BR$". For example, $\mathbf{P}_t^{BL}$ denotes a frame of 3D point clouds from 3D LiDAR at epoch $t$. The orientation provided by the AHRS is denoted as $\mathbf{R}_{BI,t}^L$ at a given epoch $t$. The extrinsic parameters between the GNSS receiver and the 3D LiDAR are denoted as $\mathbf{T}_{BL}^{BR} = [\mathbf{R}_{BL}^{BR} \quad \mathbf{t}_{BL}^{BR}]$. An illustration of the coordinate system is shown in Fig. 2.

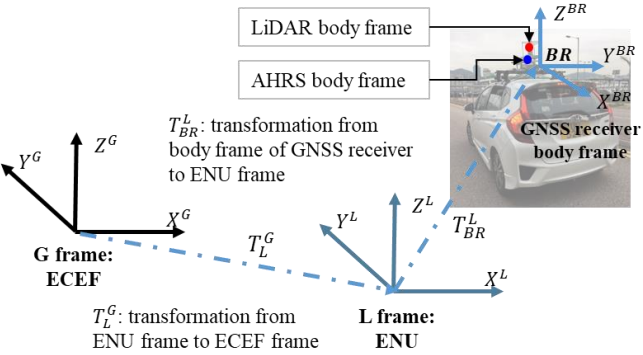

Fig. 2. Overview of the coordinate systems adopted in this paper. The ECEF frame is fixed on the center of the earth. The first point is selected as the reference of the ENU frame. The extrinsic parameters between the LiDAR, AHRS, and GNSS receiver are fixed and calibrated beforehand.

Table 1. Notions in this paper

| Notation | Description | Notation | Description |
|----------|-------------|----------|-------------|
| $t$ | GNSS epoch | "$G$" | ECEF frame |
| $s$ | Index of satellite | "$L$" | ENU frame |
| $r$ | GNSS receiver | "$BI$" | AHRS body frame |
| $\rho$ | Pseudorange | "$BL$" | LiDAR body frame |
| $\rho_{r,t}^s$ | Pseudorange of satellite $s$ at epoch $t$ | "$BR$" | GNSS receiver body frame |
| $\mathbf{p}_t^{G,s}$ | Position of satellite $s$ at epoch $t$ | $\mathbf{p}_{r,t}^G$ | Position of GNSS receiver at epoch $t$ |
| $\delta_{r,t}$ | Clock bias of GNSS receiver | $\delta_{r,t}^s$ | Clock bias of satellite |
| $\psi_{r,t}^s$ | Signal noise to the ratio (SNR) | $K_w$ | Scaling factor for weightings |
| $\varepsilon_{r,t}^s$ | Satellite elevation angle | $\alpha_{r,t}^s$ | Satellite azimuth angle |
| $k$ | Index of the search point | $N_k$ | Number of neighboring points |

## III. Sliding Window Map Generation

This section describes in detail the methodology of generating an SWM, so-called the environment description, for further NLOS detection and correction. In our previous work, described in [16, 17, 23], only real-time 3D point clouds were applied to further detect the NLOS satellites. Unfortunately, the capability of NLOS detection is limited by the FOV of 3D LiDAR. To solve this problem, we accumulate real-time 3D point clouds into a map that can effectively enhance the FOV of 3D LiDAR sensing.

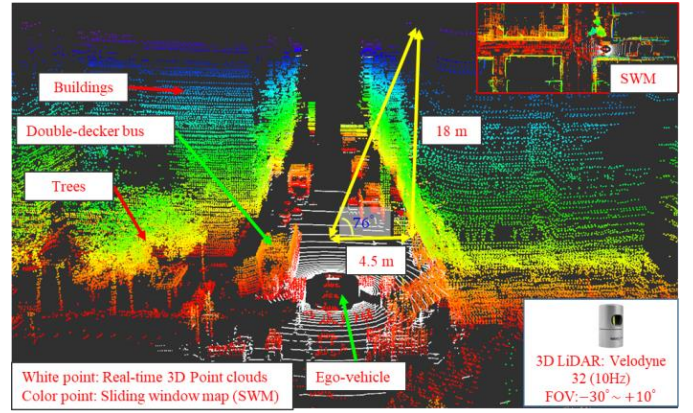

Fig. 3. Demonstration of a generated SWM and the real-time 3D point clouds. The white points represent the real-time 3D point clouds from a single frame. The colored points come from the SWM, and the color is annotated by the height information. A video illustration of the SWM generation can be found through https://www.youtube.com/watch?v=I6NTldUAAUM.

Fig. 3 shows the difference between the real-time 3D LiDAR point clouds and the SWM. The white points represent the real-time point clouds from the 3D LiDAR. It is shown from the Figure that only the low-lying parts of buildings or double-decker buses are scanned by the single-frame 3D point clouds (we used Velodyne 32 [15] in this paper). The visibility of satellites with high elevation angles cannot be effectively classified simply based on real-time 3D point clouds. Real-time 3D point clouds are also sparse, due to the physical scanning angle distribution of the 3D LiDAR sensor. However, the SWM proposed in this paper can effectively ameliorate this problem. The colored points denote the map points of the SWM. Note that the ground points were removed from the SWM for efficient NLOS detection, as described in Section IV. We can see from Fig. 3 that the elevation mask angle can reach 76° with the help of SWM, so the visibility of a satellite with an elevation angle of less than 76° can be classified in this case. The point clouds in SWM are significantly denser than raw real-time 3D point clouds, which can contribute significantly to the accuracy of NLOS detection. A snapshot of the complete SWM map is shown at the top right of Fig. 3. Both the buildings and the dynamic objects, such as double-decker buses, and even the trees are involved in the SWM, which is not included in the 3D building model [24].

To generate a point cloud map based on real-time 3D point clouds, simultaneous localization, and mapping (SLAM) [25] methods have been extensively studied over the past decades. Satisfactory accuracy can be obtained in a short period with low drift [26]. However, the error can accumulate over time, causing large errors after long-term traveling and the loop closure is not usually available. In practice, only the objects inside a circle with a radius of about 250 meters [16] can cause GNSS NLOS receptions, and buildings far away can be ignored. We, therefore, employed only the last $N_{sw}$ frames of the 3D point clouds to generate a sliding window map. As a result, the drift error for map generation is therefore bounded at a small value and is determined by the size of the window ($N_{sw}$). The LiDAR odometry and mapping (LOAM) algorithm presented in [20] is a well-known method for LiDAR odometry with outperforming accuracy in the evaluated KITTI datasets. However, according to our recent evaluation in [21], the performance of the LOAM is significantly challenged in urban canyons with numerous dynamic objects, leading to distinct drift in the vertical direction. To fill this gap, our team's recent work in [21] proposed to employ the absolute ground to constrain the vertical drift and significantly improved performance is obtained. Therefore, we generate the SWM on top of our previous work in [21]. The detail of the SWM generation is shown in Algorithm 1.

---

**Algorithm 1: SWM generation based on 3D point clouds**

---

**Inputs**: A series of point clouds from epoch $t - N_{sw} + 1$ to epoch $t$ as $\{\mathbf{P}_{t-N_{sw}+1}^{BL}, \mathbf{P}_{t-N_{sw}+2}^{BL}, ..., \mathbf{P}_t^{BL}\}$. The extrinsic parameters between 3D LiDAR, AHRS, and GNSS receiver.

**Outputs**: The SWM $\mathbf{M}_t^L$

**Step 1**: Initialize $\mathbf{M}_t^L \leftarrow$ empty.

**Step 2**: SWM generation

- **Step 2-1**: Register all the point clouds $\{\mathbf{P}_{t-N_{sw}+1}^{BL}, \mathbf{P}_{t-N_{sw}+2}^{BL}, ..., \mathbf{P}_t^{BL}\}$ into a local map ($\mathbf{M}_t^{BL}$) with the $\mathbf{P}_t^{BL}$ as the first frame, based on the method in [21].

- **Step 2-2**: A point $\mathbf{M}_{t,i}^{BL}$ inside the local map $\mathbf{M}_t^{BL}$ is transformed into the receiver body frame as $\mathbf{M}_{t,i}^{BR}$ composing the $\mathbf{M}_t^{BR}$

$$\mathbf{M}_{t,i}^{BR} = \mathbf{R}_{BL}^{BR} \mathbf{M}_{t,i}^{BL} + \mathbf{t}_{BL}^{BR} \quad (1)$$

- **Step 2-3**: A point $\mathbf{M}_{t,i}^{BR}$ inside the local map $\mathbf{M}_t^{BR}$ can be transformed into the ENU frame as $\mathbf{M}_{t,i}^L$ composing the $\mathbf{M}_t^L$

$$\mathbf{M}_{t,i}^L = \mathbf{R}_{BI,t}^L \mathbf{R}_{BR}^{BI} \mathbf{M}_{t,i}^{BL} + \mathbf{t}_{BR}^L \quad (2)$$

---

In this case, the SWM is generated as $\mathbf{M}_t^L$ represented in the ENU frame, which is to be used for GNSS NLOS detection and correction in the next section. The size $N_{sw}$ is set to 200 in this paper.

## IV. GNSS POSITIONING WITH PSEUDORANGE MEASUREMENT RECTIFICATION

Recently, the work in [27] described a novel general online sensor model validation and estimation framework. The framework consists of three phases: model validation, model calibration, and model repair. The authors proposed that sensor measurements should be validated, calibrated, or repaired before their integration with data from other sensors. The main reason behind this is that sensor measurements can be affected or polluted by environmental conditions, causing violations of the assumptions of the original sensor model. Many sensor measurements can violate the assumptions of the standard sensor model in challenging environments, such as urban canyons. For example, LiDAR-based positioning can be severely degraded in an urban canyon with numerous dynamic objects [28]. Therefore, a fixed sensor model cannot bound the potential error of LiDAR-based positioning. Therefore, the ability to effectively validate, calibrate, and repair the sensor model as required is valuable for sensor fusion in such areas.

Following the framework proposed by Jurado and Raquet [27], we applied the three phases to the GNSS pseudorange measurements. First, model validation is performed based on satellite visibility classification using SWM. Second, if one satellite is classified as NLOS, we proceed to the model calibration phase, which re-estimates the GNSS measurement by correcting the NLOS pseudorange measurement. However, if one satellite is classified as NLOS, but its reflecting point is not found inside the SWM, which means NLOS correction is not available, we proceed to the model repair phase by de-weighting the NLOS measurements for use in further positioning. The remainder of this section describes these three phases.

### A. Model Validation: NLOS Detection Based on SWM

In this section, we describe the details of NLOS detection based on the SWM. Unlike the 3D building models, which consist of consistent surfaces from buildings [24], the SWM only provides unorganized discrete points. To effectively classify satellite visibility based on the SWM, we developed a fast searching method (Algorithm 2).

The inputs of the algorithm include the SWM ($\mathbf{M}_t^L$), elevation angle $\varepsilon_{r,t}^s$ of satellite $s$, the azimuth angle $\alpha_{r,t}^s$ at epoch $t$, the maximum searching distance, $D_{thres}$, and a constant incremental value, $\Delta d_{pix}$. The output is the satellite visibility, $v_{r,t}^s$, of satellite $s$. In Step 1, a search point is initialized at the center of the 3D LiDAR denoted by $(x_t^{L,c}, y_t^{L,c}, z_t^{L,c})$ in the ENU frame. The superscript $c$ denotes the center of 3D LiDAR. A search direction connecting the GNSS receiver and the satellite is determined by the elevation and azimuth angle of satellite $s$. The SWM is transformed into a *kdTree* structure [29], $\mathbf{M}_{t,tree}^L$, for finding neighboring points. The *kdTree* is a special structure for point cloud processing which can perform efficiently when searching neighboring points. In Step 2, given a fixed incremental value, $\Delta d_{pix}$, the search point is moved to the next point ($x_{t,k}^{L,c}, y_{t,k}^{L,c}, z_{t,k}^{L,c}$) calculated using (3)-(5), based on the search direction shown on the left-hand side of Fig. 4. The $k$ denotes the index of the search point. The number ($N_k$) of neighboring points near the search point is counted. If $N_k$ exceeds a certain threshold $N_{thres}$, there are some map points from buildings or dynamic objects near the search point ($x_{t,k}^{L,c}, y_{t,k}^{L,c}, z_{t,k}^{L,c}$), and we consider that the line-of-sight connecting the GNSS receiver and satellite is blocked. Therefore, satellite $s$ is classified as an NLOS satellite. Otherwise, repeat Steps 2 and 3. If $k\Delta d_{pix} > D_{thres}$, it means that the direction between the GNSS receiver and the satellite is line-of-sight. In this work, $D_{thres}$ was set to 250 meters, so points within 250 meters were considered for NLOS detection. Only the direction connecting the GNSS receiver and the satellite needs to be considered, instead of traversing the whole SWM, an approach that contributes to the efficiency of NLOS detection.

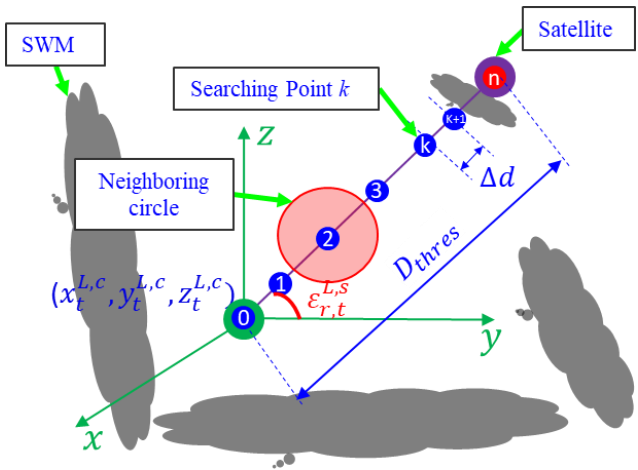

Fig. 4. Illustration of NLOS detection based on SWM.

Therefore, satellite visibility can be classified using Algorithm 2. A demonstration of the satellite visibility classification result is shown in Fig. 5. The red and blue circles represent the NLOS and LOS satellites, respectively. The length of the line connecting the center of 3D LiDAR and the satellite is $D_{thres}$. In our implementation, less than 10 ms was spent on classifying the visibility of each satellite.

The numbers near the circles in Fig. 5 denote the elevation angle of the corresponding satellite. We can see that the NLOS satellite with an elevation angle of 54 degrees was detected. As shown in Fig. 3, the maximum mask elevation angle can reach 76 degrees. In practice, the maximum mask elevation angle based on SWM was significantly correlated with the width of the street. The narrower the street was, the higher the mask elevation angle is achieved. Although the proposed SWM has effectively enhanced the FOV of LiDAR sensing compared with our previous work described in [16, 17, 23], the SWM still could not fully reconstruct scenarios with very tall buildings. However, according to recent research described in [16], NLOS satellites with low elevation angles produce the majority of GNSS positioning errors.

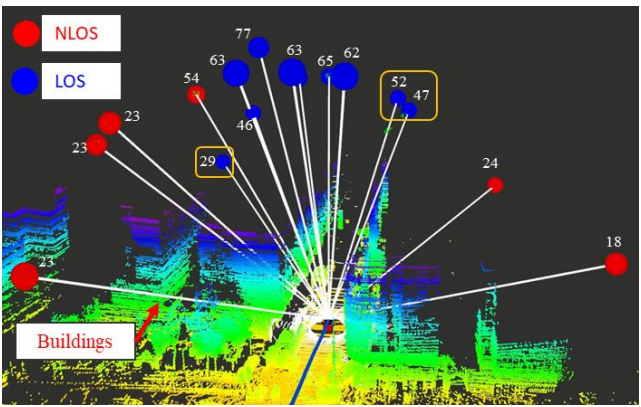

Fig. 5. Demonstration of NLOS detection based on SWM. A detailed video can be found through https://www.youtube.com/watch?v=I6NTldUAAUM. The circles annotated by the yellow rectangles are NLOS satellites that are not detected by the proposed method.

---

**Algorithm 2: NLOS Detection based on SWM**

**Inputs**: Point clouds $\mathbf{M}_t^L$, $\varepsilon_{r,t}^s$, $\alpha_{r,t}^s$.

**Outputs**: Satellite visibility $v_{r,t}^s$

**Step 1**: Initialize the searching point at $(x_t^{L,c}, y_t^{L,c}, z_t^{L,c})$, the searching direction denoted by $\varepsilon_{r,t}^s$ and $\alpha_{r,t}^s$, transform the $\mathbf{M}_t^L$ into *kdTree* and get $\mathbf{M}_{t,tree}^L$.

**Step 2**: Given a constant incremental value $\Delta d_{pix}$, the searching point is updated as follows:

$$x_{t,k}^{L,c} = x_{t,k-1}^{L,c} + \Delta d_{pix} \sin(\alpha_{r,t}^s)\cos(\varepsilon_{r,t}^s) \quad (3)$$

$$y_{t,k}^{L,c} = y_{t,k}^{L,c} + \Delta d_{pix} \cos(\alpha_{r,t}^s)\cos(\varepsilon_{r,t}^s) \quad (4)$$

$$z_{t,k}^{L,c} = z_{t,k-1}^{L,c} + \Delta d_{pix} \sin(\varepsilon_{r,t}^s) \quad (5)$$

**Step 3**: if $k\Delta d_{pix} < D_{thres}$, find the nearest neighbor points (NNPs) of a given point $(x_{t,k}^{L,c}, y_{t,k}^{L,c}, z_{t,k}^{L,c})$ and get $N_k$ NNPs.

**Step 4**: Repeat Step 2~3, until $N_k > N_{thres}$. Then the satellite is NLOS ($v_{r,t}^{L,s}=0$), else LOS ($v_{r,t}^{L,s}=1$)

---

### B. Model Calibration: NLOS Correction Based on SWM

This section presents the details of NLOS correction (model calibration) based on an SWM. To effectively estimate the potential NLOS error, the distance between the GNSS receiver, satellite elevation, and azimuth angles are needed based on the NLOS error model which is proposed in [30]. Therefore, the key is to detect the reflecting point corresponding to the NLOS. A ray-tracing [31] technique is commonly used to simulate NLOS signal transmission routes for finding the NLOS reflectors in range-based 3DMA GNSS [5, 32-34]. This approach can incur a high computational load. However, unlike the 3D building models, the SWM described in this paper does not produce continuous building surfaces and clear building boundaries. The SWM only provides large amounts of dense, discrete, unorganized point clouds; there are about 10 million points inside an SWM. Instead of applying the ray-tracing technique to find the reflectors inside the SWM, we directly search for the reflectors from the SWM, using an efficient *kdTree* structure.

The details of the reflector detection algorithm are presented in Algorithm 3. The inputs of the algorithm include the SWM ($\mathbf{M}_t^L$), elevation angle $\varepsilon_{r,t}^s$ of satellite $s$, the azimuth angle $\alpha_{r,t}^s$ at epoch $t$, and the azimuth resolution, $\alpha_{res}$, as in Algorithm 2. The output is the closest reflecting point, $\mathbf{p}_{r,t}^{L,s}$, which is the most probable reflector for an NLOS satellite $s$.

**Step1**. A search point is initialized at the center of the 3D LiDAR. The search direction is determined based on the satellite elevation, $\varepsilon_{r,t}^s$, and azimuth $\alpha_{r,t}^s$.

**Step2**. For a typical signal reflection route, two segments are included. The first segment is the signal transmission from the satellite to the reflector. The second segment is the signal transmission from the reflector to the GNSS receiver. According to our previous work in [35], the reflected signal should have the same elevation angle as the expected directed signal. Therefore, we traverse all the azimuths from 0 to 360 degrees, with an azimuth resolution of $\alpha_{res}$ and elevation angle of $\varepsilon_{r,t}^s$, to find all the possible routes of NLOS transmissions. For example, for a given direction specified by $\varepsilon_{r,t}^s$ and the azimuth angle $\alpha_s$, the line-of-sight between the GNSS receiver and satellite is identified based on Algorithm 2.

**Step3**. If the line-of-sight associated with the direction is blocked by a point, $p_j$, (Step 2 in Algorithm 3), this means that the $p_j$ can potentially be a reflecting point. Meanwhile, if the line-of-sight connecting the point $p_j$ and the satellite is not blocked (Step 3 in Algorithm 3), point $p_j$ is considered as a possible reflector and is saved to $\mathbf{Q}_{r,t}^{L,S}$.

**Step 4**. The $\alpha_s$ proceed to the next azimuth based on Step 4. By repeating steps 2 and 3, all possible reflectors are identified, based on the assumption of the same elevation angles. Fig. 5 (a)- (b) shows the result of possible reflector detection for 1-2 NLOS satellites. We observe that multiple possible reflectors are found using Steps 1 to 4. The red circles in Fig. 5 denote the NLOS satellite, and the red lines denote possible NLOS reflection and transmission routes. According to [30], the reflector with the shortest distance is usually the best candidate.

**Step 5**. A unique reflector can be detected based on the shortest distance assumption (Step 5), as shown in Fig. 5 (c). Therefore, the reflecting point for a given satellite $s$ is detected as $p_{r,t}^{L,S}$, and the distance between the GNSS receiver can be calculated accordingly.

---

**Algorithm 3: Reflecting Point Detection (RPD) based on SWM**

---

**Inputs**: Point clouds $\mathbf{M}_t^L$, $\varepsilon_{r,t}^s$, $\alpha_{r,t}^s$ and azimuth resolution as $\alpha_{res}$.

**Outputs**: Reflecting point $\mathbf{p}_{r,t}^{L,S}$.

**Step 1**: Initialize the searching point at $(x_t^{L,c}, y_t^{L,c}, z_t^{L,c})$, the searching direction denoted by $\varepsilon_{r,t}^s$ and $\alpha_{r,t}^s$, transfer the $\mathbf{M}_t^L$ into $kdTree$ and get $\mathbf{M}_{t,tree}^L$. Initialize reflecting points array $\mathbf{Q}_{r,t}^{L,S}$. $\alpha_s = 0$.

**Step 2**: Get the first point $\mathbf{p}_j$ inside the map blocking the searching direction denoted by $\varepsilon_{r,t}^s$ and $\alpha_s$ using Algorithm 2. if $\mathbf{p}_j$ is found, go to Step 3, otherwise go to Step 4.

**Step 3**: If the direction connecting the point and satellite is visible, save $\mathbf{p}_j$ to $\mathbf{Q}_{r,t}^{L,S}$, which can be a possible reflector.

**Step 4**: $\alpha_s \leftarrow (\alpha_s + \alpha_{res})$, repeat Step 1 to 2 until $\alpha_s > 360°$.

**Step 5**: find the most likely reflector $\mathbf{p}_j$ from $\mathbf{Q}_{r,t}^{L,S}$ with the shortest distance between the GNSS receiver and the reflector. $\mathbf{p}_{r,t}^{L,S} \leftarrow \mathbf{p}_j$.

---

The proposed NLOS reflector detection method does not rely on the accuracy of the detection of building surfaces. The short distance assumption applied in Step 5 of Algorithm 3 can effectively prevent overcorrection, as only the closest reflector is identified as the unique reflector. Therefore, the reflector can be detected accordingly. Then the potential NLOS delay for a satellite $s$ can be calculated as $d_{r,t}^{L,S}$ [30]:

$$d_{r,t}^{L,S} = ||\mathbf{p}_{r,t}^{L,S}|| \, sec(\varepsilon_{r,t}^s) + ||\mathbf{p}_{r,t}^{L,S}|| \, sec(\varepsilon_{r,t}^s)\cos\left(2\varepsilon_{r,t}^s\right) \quad (6)$$

where the operator $||*||$ is used to calculate the norm of a given vector. The $sec\,(*)$ denotes the secant function. Due to the sparsity of the SWM, although it is still denser than the 3D real-time point clouds, there are still some satellites whose reflectors cannot be found using the SWM. Therefore, we remodel NLOS satellites whose reflectors are not found, using the approach described in the next section.

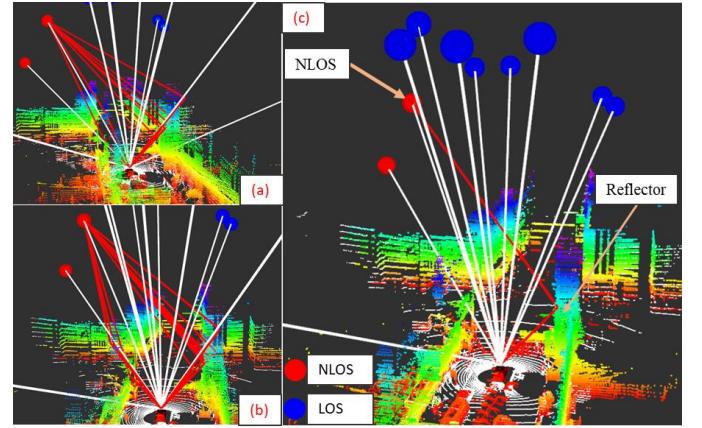

Fig. 6. Demonstration of the NLOS signal reflector detection. Red and blue circles represent NLOS and LOS satellites, respectively. White lines denote LOS transmission routes. Red lines represent NLOS transmission routes. Multiple possible NLOS transmission routes are shown in (a)-(b). The most probably NLOS transmission route is shown in (c), based on the shortest route assumption adopted in Algorithm 3.

### C. Model Repair: NLOS Remodeling

According to [30], satellites with lower elevation angles and smaller SNR have a higher possibility of contamination by NLOS errors [30]. Pseudorange uncertainty modeling based on the satellite elevation angle and SNR was reported in [36, 37]. The weighting scheme in [37] produces satisfactory performance in open areas. However, the scheme may not work in dense urban areas, as the NLOS can have high elevation angles and SNR, as can be seen in our previous work [38]. This weighting scheme treats the LOS and NLOS in the same manner, which is not preferable when the NLOS has already been detected. The weighting scheme in [36] employs a scaling factor to assign the LOS and NLOS different weightings. Inspired by this approach, this paper model the uncertainty of LOS and NLOS using the weighting scheme described in [37]. A scaling factor was added to the scheme to de-weight the NLOS measurements. The selection of the scaling factor can be found in our previous work [39]. Therefore, the weighting for each satellite is estimated as follows:

(1) if a satellite is classified as a LOS measurement, the weighting is calculated based on satellite SNR and elevation angle [37].

(2) if a satellite is classified as an NLOS measurement and the pseudorange error is corrected, the weighting is calculated based on satellite SNR and elevation angle [37].

(3) if a satellite is classified as an NLOS measurement but the reflecting point is not detected, the weighting is calculated based on satellite SNR and elevation angle together with a scaling factor $K_w$ [39].

### D. GNSS Positioning Via Weighted Least Square

The pseudorange measurement from the GNSS receiver, $\rho_{r,t}^s$, is denoted as follows [40]:

$$\rho_{r,t}^s = r_{r,t}^s + c(\delta_{r,t} - \delta_{r,t}^s) + I_{r,t}^s + T_{r,t}^s + \varepsilon_{r,t}^s \quad (7)$$

where $r_{r,t}^s$ is the geometric range between the satellite and the GNSS receiver. $I_{r,t}^s$ represents the ionospheric delay distance;

$T_{r,t}^s$ indicates the tropospheric delay distance. $\varepsilon_{r,t}^s$ represents the noise caused by the multipath effects, NLOS receptions, receiver noise, antenna phase-related noise. Meanwhile, the atmosphere effects ($T_{r,t}^s$ and $I_{r,t}^s$) are compensated using the conventional models (Saastamoinen and Klobuchar models, respectively) presented in RTKLIB [41].

The observation model for GNSS pseudorange measurement from a given satellite $s$ is represented as follows:

$$\rho_{r,t}^s = h_{r,t}^s(\mathbf{p}_{r,t}, \mathbf{p}_t^s, \delta_{r,t}) + \varepsilon_{r,t}^s \tag{8}$$

$$\text{with } h_{r,t}^s(\mathbf{p}_{r,t}^G, \mathbf{p}_t^{G,S}, \delta_{r,t}) = ||\mathbf{p}_t^{G,S} - \mathbf{p}_{r,t}^G|| + \delta_{r,t}$$

where the variable $\varepsilon_{r,t}^s$ stands for the noise associated with the $\rho_{r,t}^s$. Be noted that the NLOS error $d_{r,t}^{LoS}$ is subtracted from $\rho_{r,t}^s$ before being used in the further GNSS positioning. The Jacobian matrix $\mathbf{G}_t^G$ for the observation function $h_{r,t}^s(*)$ is denoted as follows:

$$\mathbf{G}_t^G = \begin{bmatrix} \frac{p_{t,x}^{G,1} - p_{r,t,x}^G}{\left\|\mathbf{p}_t^{G,1} - \mathbf{p}_{r,t}^G\right\|} & \frac{p_{t,y}^{G,1} - p_{r,t,y}^G}{\left\|\mathbf{p}_t^{G,1} - \mathbf{p}_{r,t}^G\right\|} & \frac{p_{t,z}^{G,1} - p_{r,t,z}^G}{\left\|\mathbf{p}_t^{G,1} - \mathbf{p}_{r,t}^G\right\|} & 1 \\ \frac{p_{t,x}^{G,2} - p_{r,t,x}^G}{\left\|\mathbf{p}_t^{G,2} - \mathbf{p}_{r,t}^G\right\|} & \frac{p_{t,y}^{G,2} - p_{r,t,y}^G}{\left\|\mathbf{p}_t^{G,2} - \mathbf{p}_{r,t}^G\right\|} & \frac{p_{t,z}^{G,2} - p_{r,t,z}^G}{\left\|\mathbf{p}_t^{G,2} - \mathbf{p}_{r,t}^G\right\|} & 1 \\ \frac{p_{t,x}^{G,3} - p_{r,t,x}^G}{\left\|\mathbf{p}_t^{G,3} - \mathbf{p}_{r,t}^G\right\|} & \frac{p_{t,y}^{G,3} - p_{r,t,y}^G}{\left\|\mathbf{p}_t^{G,3} - \mathbf{p}_{r,t}^G\right\|} & \frac{p_{t,z}^{G,3} - p_{r,t,z}^G}{\left\|\mathbf{p}_t^{G,3} - \mathbf{p}_{r,t}^G\right\|} & 1 \\ \cdots & \cdots & \cdots & \cdots \\ \frac{p_{t,x}^{G,m} - p_{r,t,x}^G}{\left\|\mathbf{p}_t^{G,m} - \mathbf{p}_{r,t}^G\right\|} & \frac{p_{t,y}^{G,m} - p_{r,t,y}^G}{\left\|\mathbf{p}_t^{G,m} - \mathbf{p}_{r,t}^G\right\|} & \frac{p_{t,z}^{G,m} - p_{r,t,z}^G}{\left\|\mathbf{p}_t^{G,m} - \mathbf{p}_{r,t}^G\right\|} & 1 \end{bmatrix} \tag{9}$$

where the variable $m$ denotes the total number of the satellite at epoch $t$. Therefore, the position of the GNSS receiver can be estimated via weighted least squares iteratively as follows:

$$\begin{bmatrix} \mathbf{p}_{r,t}^G \\ \delta_{r,t} \end{bmatrix} = (\mathbf{G}_t^{G^T} \mathbf{W}_t \mathbf{G}_t^G)^{-1} \mathbf{G}_t^{G^T} \mathbf{W}_t \begin{bmatrix} \rho_{r,t}^1 \\ \vdots \\ \rho_{r,t}^m \end{bmatrix} \tag{10}$$

where the $\mathbf{W}_t$ denotes the weighting matrix based on the weightings estimated in Section IV-C as follows:

$$\mathbf{W}_t = \begin{bmatrix} \omega_{r,t}^1 & 0 & 0 & 0 \\ 0 & \omega_{r,t}^2 & 0 & 0 \\ 0 & 0 & \ddots & 0 \\ 0 & 0 & 0 & \omega_{r,t}^s \end{bmatrix} \tag{11}$$

$$\text{With } \omega_{r,t}^s = \begin{cases} \text{f}(\varepsilon_{r,t}^s, \psi_{r,t}^s), & \text{if LOS or CNLOS} \\ K_w \, \text{f}(\varepsilon_{r,t}^s, \psi_{r,t}^s), & \text{if FNLOS} \end{cases}$$

where the function $\text{f}(\varepsilon_{r,t}^s, \psi_{r,t}^s)$ is defined to calculate the weighting of the LOS measurements based on satellite SNR and elevation angle as follows

$$\text{f}(\varepsilon_{r,t}^s, \psi_{r,t}^s) = = \frac{1}{\sin^2 \varepsilon_{r,t}^s} \left( 10^{-\left(\frac{\psi_{r,t}^s - \tau}{a}\right)} \left( \left( \frac{A}{10^{-\left(\frac{(\tau-\tau)}{a}\right)}} - 1 \right) \frac{(\psi_{r,t}^s - \tau)}{F - T} + 1 \right) \right) \tag{12}$$

where $T$ indicates the SNR threshold and the parameters $a$, $A$, and $F$ are selected based on [37].

## V. EXPERIMENT RESULTS AND DISCUSSION

### A. Introduction of the Experiment

**Experimental scenes**: To verify the effectiveness of the proposed method, two experiments were conducted in typical urban canyons in Hong Kong (Fig. 7). The left and right figures show the scene of the evaluated urban canyons. Both of the urban scenarios contain static buildings, trees, and dynamic objects, such as double-decker buses. We are aware of the limitation of the method described in this paper, mentioned at the end of Section IV-A, that the sliding window map cannot sense the roof of buildings above 40 meters high in dense urban environments. We first experimented with a typical urban canyon in Hong Kong (urban canyon 1 in Fig. 7). Then we performed another experiment in a highly urbanized area in Hong Kong (urban canyon 2 in Fig. 7), where the buildings are significantly higher, and which is one of the densest areas in Hong Kong, to study the limitations of the proposed method. Some NLOS satellites reflected by buildings taller than 40 meters may not be detected using SWM in urban canyon 2.

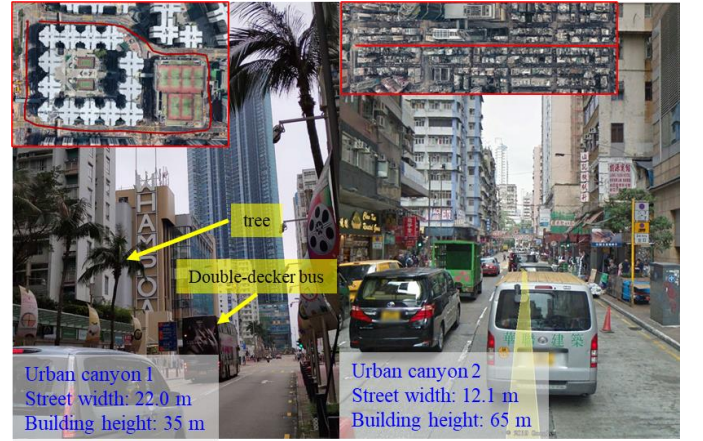

Fig. 7. Demonstration of the evaluated urban canyons 1 and 2.

**Sensor setups**: The detail of the data collection vehicle can be found through our open-sourced UrbanNav dataset [2] [1]. In both experiments, a u-blox M8T GNSS receiver was used to collect raw GNSS measurements at a frequency of 1 Hz. A 3D LiDAR sensor (Velodyne 32) was employed to collect raw 3D point clouds at a frequency of 10 Hz. The Xsens Ti-10 INS was employed to collect data at a frequency of 100 Hz. Besides, the NovAtel SPAN-CPT, a GNSS (GPS, GLONASS, and Beidou) RTK/INS (fiber-optic gyroscopes, FOG) integrated navigation system was used to provide ground truth of positioning. The gyro bias in-run stability of the FOG is 1 degree per hour, and its random walk is 0.067 degrees per hour. The baseline between the rover and the GNSS base station is about 7 km. All the data were collected and synchronized using a robot operation system (ROS) [42]. The coordinate systems between all the sensors were calibrated before the experiments.

**Evaluated methods**: We analyzed the performance of 3D LiDAR aided GNSS positioning by comparing five methods, as shown below to validate the effectiveness of the proposed method in improving the GNSS positioning.

(a) **u-blox**: the GNSS positioning solution from the u-blox

M8T receiver.

(b) **WLS**: weighted least squares (WLS) method [37].

(c) **WLS-NE**: weighted least squares (WLS) method [37] with all NLOS satellites excluded.

(d) **R-WLS**: WLS method with the aid of the re-weighting scheme with all NLOS satellites being re-weighted.

(e) **CR-WLS** (proposed SPP): WLS method with the aid of (1) the NLOS correction proposed in Section IV-B if the reflector was detected, and (2) the re-weighting scheme in Section IV-C if the reflector was not detected.

### B. Positioning Performance Evaluation

#### 1) GNSS Positioning in Urban Canyon 1

The results of the GNSS positioning experiments using the five methods are shown in Table 2. The first column shows the 2D positioning error of the u-blox receiver. The positioning result is based on standard NMEA [22] messages from the u-blox receiver. A mean error of 31.02 meters was obtained, with a standard deviation of 37.69 meters. The maximum error reached 177.59 meters due to the severe multipath and NLOS receptions from surrounding buildings. The GNSS solution was available throughout the experiment (100% availability). The second column shows the positioning result using the raw pseudorange measurements from the u-blox receiver and positioning based on WLS. Similarly, the weighting scheme was taken from [37] and is based on the satellite elevation angle and the signal-to-noise ratio (SNR). The positioning error decreased to 9.57 meters with a standard deviation of 7.32 meters. The maximum error also decreased to less than 50 meters. The positioning error increased to 11.63 meters after excluding all detected NLOS satellites, a result that is even worse than that of the WLS. This situation arose because excessive NLOS exclusion can significantly distort the perceived geometric distribution of the satellites. Our previous results, described in [16, 38], showed a similar phenomenon. The standard deviation also increased compared with that of the WLS. Availability decreased slightly from 100% (WLS) to 96.01%. Complete NLOS exclusion is therefore not preferable in dense urban canyons. The fourth column of the table presents the results of R-WLS where all the NLOS satellites were remodeled based on the weighting scheme described in Section 4.3, instead of excluding the NLOS satellites detected. The 2D mean error was reduced from 9.57 meters (WLS) to 9.01 meters. Both the standard deviation and maximum errors decreased slightly. The last column shows the 2D positioning error of CR-WLS. The 2D positioning error decreased to 7.92 meters, with a standard deviation of 5.27 meters. Availability is also guaranteed using the proposed method (CR-WLS). The improved GNSS positioning results demonstrate the effectiveness of the proposed method in mitigating the effects of NLOS signals.

In short, the best performance of GNSS positioning was obtained using CR-WLS. These improved results show that the proposed method can mitigate the effects of NLOS receptions by remodeling and correcting NLOS signals. Due to the complementarity of GNSS and INS, it is interesting to see how the remodeling and correction of GNSS measurements can contribute to the GNSS/INS integration which is to be verified in the next section.

Table 2. Positioning performance of GNSS SPP in urban canyon 1 (Max: maximum error, Avail: availability of GNSS solution)

| All Data | u-blox | WLS | WLS-NE | R-WLS | CR-WLS |
|---|---|---|---|---|---|
| **MEAN (m)** | 31.02 | 9.57 | 11.63 | 9.01 | 7.92 |
| **STD (m)** | 37.69 | 7.32 | 13.05 | 6.90 | 5.27 |
| **Max (m)** | 177.59 | 46.29 | 52.93 | 43.59 | 41.75 |
| **Avail** | 100% | 100% | 96.01% | 100% | 100% |

#### 2) GNSS/INS Integrated Positioning in Urban Canyon 1

In this section, we present the results of GNSS/INS integration. In this evaluation, Three GNSS/INS integrated positioning methods were also compared:

(1) **EKF**: standard EKF-based tightly coupled GNSS/INS integration based on [43].

(2) **FG**: factor graph-based tightly-coupled GNSS/INS integration [44].

(3) **FG-3DLA** (proposed integration): factor graph-based tightly coupled GNSS/INS integration with the help of 3D LiDAR aided GNSS pseudorange measurement rectification. The implementation of GNSS/INS integration using factor graph optimization (FGO) is based on our recent work in [44].

Table 3 shows the 2D positioning errors obtained using the three kinds of GNSS/INS integrations. A 2D mean error of 8.03 meters was obtained using EKF, with a maximum error of 44.55 meters. Significantly improved positioning accuracy was obtained after the application of the state-of-the-art FGO, with the mean error decreasing from 8.03 to 3.64 meters. Both the standard deviation and the maximum error decreased.

Our recent research, described in [44], extensively evaluated the performance of GNSS/INS integration using EKF and factor graphs. Unlike conventional EKF based GNSS/INS integration, the FGO makes use of historical measurements during optimization, which exploits the connectivity between historical states and measurements. Unfortunately, the improvements from the FGO are still limited if the GNSS measurements are not well modeled. The maximum error still reached 23.56 meters. The positioning error still fluctuates dramatically between epochs 190 and 205 (see Fig. 8). The main reason for this variability is the high number of unmodeled NLOS measurements.

Table 3. Positioning performance of GNSS/INS Integration in urban canyon 1

| GNSS/INS | EKF | FG | FG-3DLA |
|---|---|---|---|
| **MEAN (m)** | 8.03 | 3.64 | 2.80 |
| **STD (m)** | 7.60 | 3.19 | 1.62 |
| **Max (m)** | 44.55 | 23.56 | 9.71 |

With the help of the proposed method, the 2D mean error decreased from 3.64 meters (FG) to 2.8 (FG-3DLA) meters. The standard deviation was also reduced to 1.62 meters. The maximum 2D error was reduced from 23.56 meters (FG) to 9.71 meters. Fig. 8 and Fig. 9 show the positioning error and trajectories, respectively. These improved results show the effectiveness of the proposed method. Although GNSS positioning using the proposed CR-WLS still reaches 7.92 meters, GNSS/INS integration using FGO can effectively make the best use of the pseudorange correction (model calibration in Section IV-B) and uncertainty modeling (model repair in Section IV-C). After applying the 3D LiDAR aided GNSS positioning, the performance of GNSS/INS integration using the state-of-the-art factor graph was pushed significantly higher.

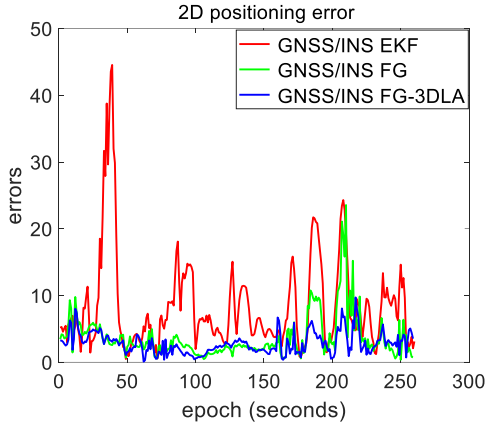

Fig. 8.    2D positioning errors of the GNSS/INS integrations in urban canyon 1

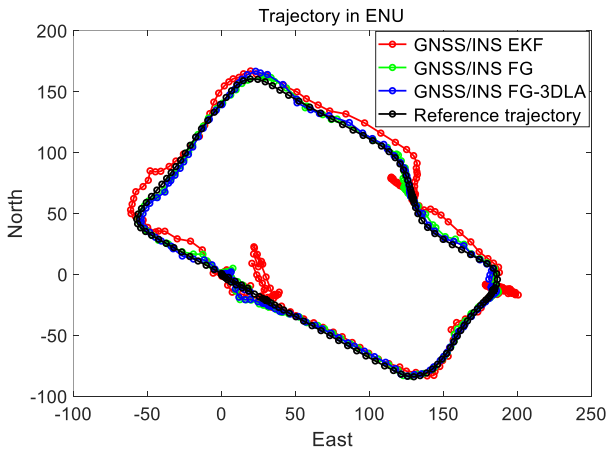

Fig. 9.    2D positioning trajectories of the GNSS/INS integrations in urban canyon 1

### 3) GNSS Positioning in Urban Canyon 2

To challenge the performance of the proposed method, another experiment was conducted in a denser urban canyon 2. As a result, NLOS satellites with high elevation angles cannot be fully detected using the SWM. We also wanted to explore what would happen in a denser urban canyon, using the proposed method.

Similar to experiment 1, the results of the GNSS positioning experiment are presented in Table 4 to show the effectiveness of the proposed method in GNSS positioning. A positioning error of 30.68 meters was obtained using the u-blox receiver with a maximum error of 92.32 meters. A GNSS positioning error of 23.79 meters was obtained using WLS based on the raw pseudorange measurements from the u-blox receiver. The maximum error increased slightly to 104.83 meters, compared with the GNSS positioning using data directly from the u-blox receiver. After excluding all detected NLOS satellites from the GNSS positioning (WLS-NE), both the mean and standard deviation increased to 25.14 and 23.73 meters, respectively. The availability of GNSS positioning data decreased to 95.52%, due to the lack of satellites for GNSS positioning, which again shows that complete NLOS exclusion is not preferable in urban canyons. With the help of NLOS remodeling, the 2D error decreased to 19.61 meters using R-WLS. One hundred percent availability is guaranteed. The GNSS positioning error was further decreased to 17.09 meters using the CR-WLS method. The improvement in the results shows the effectiveness of the proposed method for 3D LiDAR aided GNSS positioning. The maximum error still reached 71.28 meters, because not all NLOS satellites can be detected and mitigated.

Table 4. Positioning performance of GNSS SPP in urban canyon 2

| All Data | u-blox | WLS | WLS-NE | R-WLS | CR-WLS |
|---|---|---|---|---|---|
| MEAN (m) | 30.68 | 23.79 | 25.14 | 19.61 | 17.09 |
| STD (m) | 26.53 | 18.22 | 23.73 | 19.80 | 20.95 |
| Max (m) | 92.32 | 104.83 | 109.30 | 86.14 | 71.28 |
| Avail | 100% | 100% | 95.52% | 100% | 100% |

### C.  Discussion: Performance and the Limitation of the Proposed LOS/NLOS Detection Using SWM

#### 1)  GNSS LOS/NLOS Detection in Urban Canyon 1

Table 5 depicts the accuracy of NLOS satellite detection. As mentioned in Section IV-A, the proposed SWM cannot fully construct all environments, so some NLOS satellites with high elevation angles cannot be detected. Therefore, we evaluated the NLOS detection performance at three elevation angle ranges. The second row in Table 5 shows the percentage of NLOS satellites that belonged to a certain elevation angle range. The NLOS satellites with elevation angles between 0° and 30° made up 43.8% of all NLOS satellites. Of these NLOS satellites, 92% were detected using the method. The NLOS detection accuracy for NLOS satellites (28.9%) with elevation angles between 30° and 60° was 35%. Similar NLOS detection accuracy (27.35%) was obtained for NLOS satellites with elevation angles between 60° and 90°. Although the NLOS satellites with high elevation angles were not detected effectively, the proposed method is a new and general solution for NLOS detection. Due to the decreased cost of 3D LiDARs, multiple 3D LiDARs [45] could be a common sensor setup for safety-critical ADV, to guarantee robustness. The use of multiple 3D LiDARs can significantly enhance the FOV of the proposed SWM. Therefore, even NLOS satellites with high elevation angles can be detected by autonomous driving vehicles using multiple 3D LiDARs [45].

Table 5. Performance of NLOS classification between different elevation ranges in urban canyon 1 (Per: percentage)

| All data | Elevation (0°–30°) | Elevation (30°–60°) | Elevation (60°–90°) |
|---|---|---|---|
| Per of NLOS Satellites | 43.8 % | 28.9% | 27.35 % |
| Accuracy of NLOS Detection | 92 % | 35 % | 21% |

#### 2)  GNSS LOS/NLOS Detection in Urban Canyon 2

To further investigate the NLOS detection performance of the proposed method in urban canyon 2, we also examined the percentage of NLOS satellites within certain elevation angle ranges in Table 6. The percentage trend is almost opposite the trend in experiment 1. The majority (44%) of the NLOS satellites belonged to the 60°–90° group in experiment 2. However, the majority of NLOS satellites belonged to the 0°–30° group in experiment 1. In experiment 2 the buildings were higher, and the streets narrower, than in experiment 1. The different trend is mainly caused by the geometry of the environments and satellite distribution. The NLOS detection accuracy for the satellites in the low elevation angle group (0°–30°) was still more than 90%, similar to experiment 1. The NLOS detection accuracy for the satellites of the high elevation angle group (60°–90°) was limited (12%). However, the proposed method can easily be adapted to ADV with multiple 3D LiDARs to further detect NLOS satellites with high elevation angles.

Since the proposed NLOS detection method relies on the orientation from the AHRS, we also present the effect of orientation error on NLOS detection accuracy. The last row of Table 6 shows the accuracy of NLOS detection at different angle ranges, using the ground truth orientation provided by the SPAN-CPT. We can see that NLOS detection accuracy increased slightly.

Table 6. Performance of NLOS classification between different elevation ranges in urban canyon 2

| All data | Elevation (0°–30°) | Elevation (30°–60°) | Elevation (60°–90°) |
|---|---|---|---|
| **Per of NLOS Satellites** | 17.7 % | 38.3% | 44.0 % |
| **Accuracy of NLOS Detection (Xsens)** | 90.7 % | 46.0% | 12.0% |
| **Accuracy of NLOS Detection (SPAN-CPT)** | 91.3 % | 47.1% | 12.5% |

### D. Discussion: Performance and the Limitation of the Proposed NLOS Correction

#### 1) GNSS LOS/NLOS Correction in Urban Canyon 1

Tables 7 and 8 show the values of NLOS correction using the proposed method in two selected epochs. Fig. 10 shows the corresponding Skyplots. In Table 7, NLOS satellite 8, with an elevation angle of 23.49° and C/N₀ of 15 dB-Hz, was detected and the NLOS correction was 10.08 meters. The fourth column shows the exact NLOS delay, labeled using a ray-tracing technique based on ground truth positioning provided by the reference system SPAN-CPT. It shows that the exact NLOS delay (15.55 meters) was slightly larger than the NLOS delay estimated using the proposed method. The major reason for the difference is that the proposed method finds the reflectors based on the shortest distance assumption. Therefore, the reflector may not be perfectly detected. In general, the results show that NLOS satellites with lower elevation angles usually cause larger NLOS delay, as shown in column four of Table 7.

The other epoch data shown in Table 8 show a slightly different trend. Satellite 30, with an elevation angle of 56.22 degrees, caused the largest NLOS delay, of 48.52 meters. According to [30], the NLOS delay is determined by the elevation angle and the distance between the GNSS receiver and the reflector. The main reason for the large NLOS delay caused by satellite 30 is the long distance between the GNSS receiver and the reflector.

Fig. 11 shows a case in which the NLOS satellites were blocked by a traffic signal pole instead of 3D buildings. Conventionally, the 3DMA GNSS only considers static buildings. However, infrastructure such as traffic signal poles and even guard bars can also cause NLOS receptions. Satellites 7, 99, and 112 were all blocked by the signal pole. With increased complexity and density of infrastructure [46], which is not included in conventional 3D building models, NLOS receptions caused by such structures should also be considered. We believe that this is also a significant contribution of the proposed method.

Table 7. NLOS pseudorange correction in urban canyon 1 at Epoch 33661 (PRN: Pseudorandom Noise Code, Ele: elevation angle, Actual: actual NLOS delay, Estimated: NLOS delay estimated by the proposed method)

| PRN | Ele (degree) | C/N₀ (dB-Hz) | Actual | Estimated |
|---|---|---|---|---|
| 8 | 23.49 | 15 | 15.55 m | 10.08 m |
| 17 | 23.13 | 18 | 13.73 m | 8.14 m |
| 11 | 62.45 | 24 | 3.87 m | 7.59 m |

Table 8. NLOS pseudorange correction in urban canyon 1 at Epoch 33730

| PRN | Ele (degree) | C/N₀ (dB-Hz) | Actual | Estimated |
|---|---|---|---|---|
| 22 | 26.91 | 19 | 12.02 m | 10.17 m |
| 28 | 28.60 | 18 | 16.41 m | 9.47 m |
| 30 | 56.22 | 30 | 48.52 m | 27.31 m |

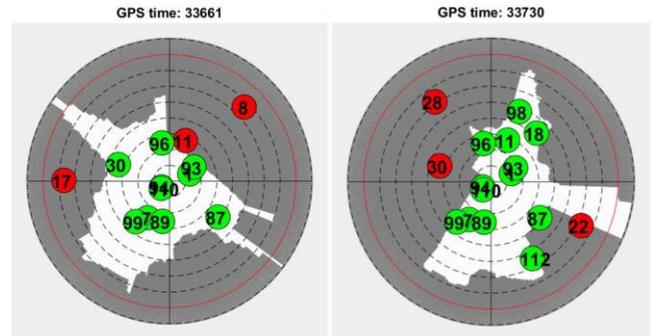

Fig. 10. Illustration of the Skyplots for epoch 33661 and 33730. The red and green circles denote the NLOS and LOS satellites, respectively. The number inside the circle represents the satellite PRN.

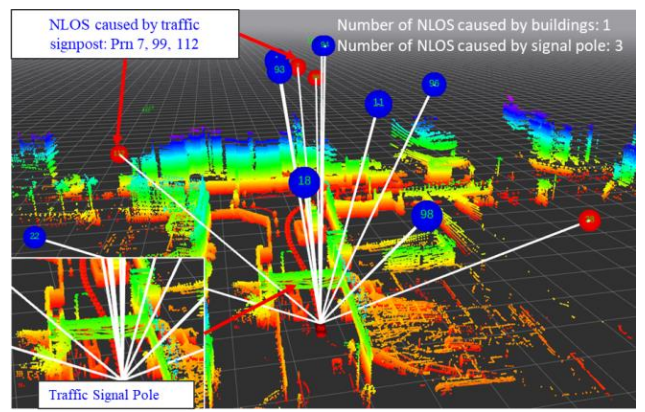

Fig. 11. Illustration of NLOS receptions blocked by an overhead traffic signal pole instead of 3D buildings. The blue and red circles denote the LOS and

NLOS satellites, respectively. The numbers inside the circles denote the satellite PRN

### 2) GNSS LOS/NLOS Correction in Urban Canyon 2

Although the mean positioning error was significantly improved compared with the 30.68 meters obtained using u-blox, it still reached 13.32 meters. The remaining error arises from two major sources: 1) undetected NLOS satellites; and 2) unexpected multipath effects. Table 8 shows the pseudorange errors caused by the multipath effects and NLOS. Satellite 15 introduced the maximum pseudorange error of 37.92 meters among the six satellites. Multipath effects can also cause errors of similar magnitude; for example, satellite 21 had a pseudorange error of 34.88 meters. Therefore, unmodeled multipath is a major factor causing the remaining 13.32 meters of positioning error. Fortunately, the multipath can be further mitigated using a higher-level GNSS antenna, which is acceptable for autonomous driving vehicles.

Table 9. Pseudorange errors in urban canyon 2 (Epoch 401793)

| PRN | Ele (degree) | C/N₀ (dB-Hz) | Type | Pseudorange Error |
|-----|--------------|--------------|------|-------------------|
| 15  | 51.6  | 31 | NLOS      | 37.29 m |
| 21  | 48.70 | 26 | Multipath | 34.88 m |
| 89  | 63.1  | 27 | NLOS      | 5.71 m  |
| 92  | 61.63 | 33 | Multipath | 5.49 m  |
| 94  | 62.32 | 32 | Multipath | 5.14 m  |
| 102 | 60.98 | 34 | Multipath | 7.77 m  |

### E. Discussion: Impacts of the size of the SWM on the Performance of the GNSS NLOS Detection

As mentioned in Section III, the window size of the SWM is determined by $N_{sw}$. The too-large window size can lead to unnecessary computation load in GNSS NLOS detection. The too-small window size can not guarantee the FOV of the SWM for the GNSS NLOS detection. To this end, this paper presents the performance of GNSS NLOS detection under different $N_{sw}$ in Table 10 based on the dataset collected in urban canyon 2. Be noted that the default value of $N_{sw}$ is 200 in this paper.

Table 10. Performance of NLOS classification between different elevation ranges in urban canyon 2 using different sizes ($N_{sw}$) of the SWM

| $N_{sw}$ | Elevation (0°–30°) | Elevation (30°–60°) | Elevation (60°–90°) |
|----------|--------------------|---------------------|---------------------|
| 100 | 50.3 % | 27.0% | 0.0%  |
| 150 | 79.3 % | 35.0% | 8.4%  |
| 200 | 90.7 % | 46.0% | 12.0% |
| 250 | 91.2 % | 46.8% | 12.0% |

Overall, it is shown in Table 10 that the performance of the GNSS NLOS detection is improved with increased $N_{sw}$. Specifically, given a value of 100 for $N_{sw}$, only 50.3% of the GNSS NLOS receptions are detected with the elevation angles ranging from 0°–30°. Meanwhile, no NLOS satellite with the elevation angles ranging from 60°–90° is detected due to the small window size of the SWM. Interestingly, when the window size $N_{sw}$ increases to 250, the improvement in GNSS NLOS detection is limited. This is because the too far away structures may not cause the signal blockage or reflection anymore.

## VI. Conclusions

GNSS positioning is currently still the major source of globally referenced positioning for intelligent transportation systems (ITS). However, accurate GNSS positioning in urban canyons is still a challenging problem. NLOS receptions currently remain the major problem for GNSS positioning in urban canyons. Therefore, effectively identifying and mitigating the effects of NLOS receptions is a significant step in achieving and popularizing accurate GNSS positioning solutions, such as SPP, real-time kinematic (RTK), and precise point positioning, in urban canyons. Since the performance of GNSS positioning relies heavily on environmental conditions, the state-of-the-art range-based 3DMA method proposed to effectively mitigate NLOS effects, based on offline environment descriptions known as 3D building models. However, with the increasing complexity and dynamics of city infrastructures, 3D building models cannot fully describe the real-time environment. Reconstructing the real-time environment based on onboard sensing is a promising method for identifying potentially polluted GNSS signals. Unlike the state-of-the-art 3DMA GNSS method, this paper proposes a novel 3D LiDAR aided GNSS positioning method which makes use of an onboard 3D LiDAR sensor to reconstruct the surrounding environment. Potential NLOS receptions caused by static buildings, dynamic objects, and even semi-static infrastructure (traffic signpost in Fig. 11) can be detected, remodeled, and even corrected. This paper reports a continuation of the previous work described in [16, 17, 23]. Three drawbacks listed in Section I are alleviated, and a general solution is proposed to mitigate the effects of NLOS receptions. The method proposed in this paper can be easily adapted to the systems with multiple 3D LiDARs, and NLOS satellites with high elevation angles can be detected accordingly. Overall, we believe that the proposed method can have a positive impact on both the academic and industrial fields.

Since the remaining positioning error still reaches about 17 meters in the dense Urban Canyon 2, we will integrate the LiDAR odometry with the proposed method in future work. The accuracy of the orientation could also be enhanced with the use of LiDAR odometry to further improve the performance of NLOS detection. The multiple LiDARs will be investigated to increase the FOV of the SWM generation, to, therefore, improve the detection and correction of GNSS NLOS with high elevation angles. Moreover, this paper only considers the single reflection from GNSS NLOS, and the multipath effects are not considered which are the key problems that limit the overall accuracy of the proposed method. In the future, we will investigate the multipath mitigation using the SWM and the multiple signal reflections will also be investigated.

## References

[1] S.-H. Kong, "Statistical analysis of urban GPS multipaths and pseudo-range measurement errors," *IEEE transactions on aerospace and electronic systems,* vol. 47, no. 2, pp. 1101-1113, 2011.

 

[2] L.-T. Hsu *et al.*, "UrbanNav: An open-sourced multisensory dataset for benchmarking positioning algorithms designed for urban areas," in *Proceedings of the 34th International Technical Meeting of the Satellite Division of The Institute of Navigation (ION GNSS+ 2021)*, 2021, pp. 226-256.

[3] W. Wen *et al.*, "Urbanloco: a full sensor suite dataset for mapping and localization in urban scenes," in *2020 IEEE International Conference on Robotics and Automation (ICRA)*, 2020, pp. 2310-2316: IEEE.

[4] P. K. Enge, "The global positioning system: Signals, measurements, and performance," *International Journal of Wireless Information Networks*, vol. 1, no. 2, pp. 83-105, 1994.

[5] S. Miura, L.-T. Hsu, F. Chen, and S. Kamijo, "GPS error correction with pseudorange evaluation using three-dimensional maps," *IEEE Transactions on Intelligent Transportation Systems*, vol. 16, no. 6, pp. 3104-3115, 2015.

[6] W. Wen, G. Zhang, and L.-T. Hsu, "GNSS NLOS exclusion based on dynamic object detection using LiDAR point cloud," *IEEE Transactions on Intelligent Transportation Systems*, 2019.

[7] J. Breßler, P. Reisdorf, M. Obst, and G. Wanielik, "GNSS positioning in non-line-of-sight context—A survey," in *Intelligent Transportation Systems (ITSC), 2016 IEEE 19th International Conference on*, 2016, pp. 1147-1154: IEEE.

[8] C. Pinana-Diaz, R. Toledo-Moreo, D. Betaille, and A. F. Gomez-Skarmeta, "GPS multipath detection and exclusion with elevation-enhanced maps," in *Intelligent transportation systems (ITSC), 2011 14th International IEEE Conference on*, 2011, pp. 19-24: IEEE.

[9] S. Peyraud *et al.*, "About non-line-of-sight satellite detection and exclusion in a 3D map-aided localization algorithm," *Sensors*, vol. 13, no. 1, pp. 829-847, 2013.

[10] N. Kbayer and M. Sahmoudi, "Performances analysis of GNSS NLOS bias correction in urban environment using a three-dimensional city model and GNSS simulator," *IEEE Transactions on Aerospace and Electronic Systems*, vol. 54, no. 4, pp. 1799-1814, 2018.

[11] L. Wang, P. D. Groves, and M. K. Ziebart, "Urban positioning on a smartphone: Real-time shadow matching using GNSS and 3D city models," 2013: The Institute of Navigation.

[12] L. Wang, P. D. Groves, and M. K. Ziebart, "GNSS shadow matching: Improving urban positioning accuracy using a 3D city model with optimized visibility scoring scheme," *Navigation*, vol. 60, no. 3, pp. 195-207, 2013.

[13] L. Wang, P. D. Groves, and M. K. Ziebart, "Smartphone shadow matching for better cross-street GNSS positioning in urban environments," *The Journal of Navigation*, vol. 68, no. 3, pp. 411-433, 2015.

[14] W. Wen and L.-T. Hsu, "3D LiDAR Aided GNSS and Its Tightly Coupled Integration with INS Via Factor Graph Optimization," presented at the ION GNSS+, 2020, Richmond Heights, Missouri, USA., 2020.

[15] F. Moosmann and C. Stiller, "Velodyne slam," in *2011 IEEE Intelligent Vehicles Symposium (IV)*, 2011, pp. 393-398: IEEE.

[16] W. Wen, G. Zhang, and L. T. Hsu, "Correcting NLOS by 3D LiDAR and building height to improve GNSS single point positioning," *Navigation*, vol. 66, no. 4, pp. 705-718, 2019.

[17] X. Bai, W. Wen, and L.-T. Hsu, "Using Sky-pointing fish-eye camera and LiDAR to aid GNSS single-point positioning in urban canyons," *IET Intelligent Transport Systems*, vol. 14, no. 8, pp. 908-914, 2020.

[18] J. S. Sánchez, A. Gerhmann, P. Thevenon, P. Brocard, A. B. Afia, and O. Julien, "Use of a FishEye camera for GNSS NLOS exclusion and characterization in urban environments," in *ION ITM 2016, International Technical Meeting*, 2016: ION.

[19] S. Kato, M. Kitamura, T. Suzuki, and Y. Amano, "Nlos satellite detection using a fish-eye camera for improving gnss positioning accuracy in urban area," *Journal of robotics and mechatronics*, vol. 28, no. 1, pp. 31-39, 2016.

[20] J. Zhang and S. Singh, "Low-drift and real-time lidar odometry and mapping," *Autonomous Robots*, vol. 41, no. 2, pp. 401-416, 2017.

[21] W. Wen, X. Bai, and L. T. Hsu, "AGPC-SLAM: Absolute Ground Plane Constrained 3D LiDAR SLAM (accepted)," *Navigation*, 2020.

[22] P. D. Groves, *Principles of GNSS, inertial, and multisensor integrated navigation systems*. Artech house, 2013.

[23] W. Wen, G. Zhang, and L.-T. Hsu, "Exclusion of GNSS NLOS receptions caused by dynamic objects in heavy traffic urban scenarios using real-time 3D point cloud: An approach without 3D maps," in *Position, Location and Navigation Symposium (PLANS), 2018 IEEE/ION*, 2018, pp. 158-165: IEEE.

[24] L.-T. Hsu, Y. Gu, and S. Kamijo, "NLOS correction/exclusion for GNSS measurement using RAIM and city building models," *Sensors*, vol. 15, no. 7, pp. 17329-17349, 2015.

[25] T. Bailey and H. Durrant-Whyte, "Simultaneous localization and mapping (SLAM): Part II," *IEEE robotics & automation magazine*, vol. 13, no. 3, pp. 108-117, 2006.

[26] J. Zhang and S. Singh, "LOAM: Lidar Odometry and Mapping in Real-time," in *Robotics: Science and Systems*, 2014, vol. 2, p. 9.

[27] J. D. Jurado and J. F. Raquet, "Towards an online sensor model validation and estimation framework," in *2018 IEEE/ION Position, Location and Navigation Symposium (PLANS)*, 2018, pp. 1319-1325: IEEE.

[28] W. Wen, L.-T. Hsu, and G. J. S. Zhang, "Performance analysis of NDT-based graph SLAM for autonomous vehicle in diverse typical driving scenarios of Hong Kong," vol. 18, no. 11, p. 3928, 2018.

[29] M. Greenspan and M. Yurick, "Approximate kd tree search for efficient ICP," in *Fourth International Conference on 3-D Digital Imaging and Modeling,*

 

2003. 3DIM 2003. Proceedings.*, 2003, pp. 442-448: IEEE.

[30] L.-T. Hsu, "Analysis and modeling GPS NLOS effect in highly urbanized area," *GPS solutions,* vol. 22, no. 1, p. 7, 2018.

[31] A. S. Glassner, *An introduction to ray tracing*. Elsevier, 1989.

[32] L.-T. Hsu, Y. Gu, and S. Kamijo, "3D building model-based pedestrian positioning method using GPS/GLONASS/QZSS and its reliability calculation," (in English), *GPS Solutions,* vol. 20, no. 3, pp. 413–428, 2016.

[33] M. Obst, S. Bauer, P. Reisdorf, and G. Wanielik, "Multipath detection with 3D digital maps for robust multi-constellation GNSS/INS vehicle localization in urban areas," in *Intelligent Vehicles Symposium (IV), 2012 IEEE,* 2012, pp. 184-190: IEEE.

[34] T. Suzuki and N. Kubo, "Correcting GNSS multipath errors using a 3D surface model and particle filter," *Proc. ION GNSS+ 2013,* 2013.

[35] H.-F. Ng, Zhang, G., Hsu, Li-Ta, "GNSS NLOS Pseudorange Correction based on Skymask for Smartphone Applications," presented at the ION GNSS+, 2019, Miami, Florida, USA, 2019.

[36] S. Tay and J. Marais, "Weighting models for GPS Pseudorange observations for land transportation in urban canyons," in *6th European Workshop on GNSS Signals and Signal Processing,* 2013, p. 4p.

[37] A. M. Herrera, H. F. Suhandri, E. Realini, M. Reguzzoni, and M. C. de Lacy, "goGPS: open-source MATLAB software," *GPS solutions,* vol. 20, no. 3, pp. 595-603, 2016.

[38] X. Bai, W. Wen, and L.-T. Hsu, "Using Sky-pointing fish-eye camera and LiDAR to aid GNSS single-point positioning in urban canyons," *IET Intelligent Transport Systems,* 2020.

[39] W. Wen, X. Bai, Y.-C. Kan, and L.-T. Hsu, "Tightly Coupled GNSS/INS Integration Via Factor Graph and Aided by Fish-eye Camera," *IEEE Transactions on Vehicular Technology,* 2019.

[40] E. Kaplan and C. Hegarty, *Understanding GPS: principles and applications*. Artech house, 2005.

[41] T. Takasu and A. Yasuda, "Development of the low-cost RTK-GPS receiver with an open source program package RTKLIB," in *International symposium on GPS/GNSS,* 2009, pp. 4-6: International Convention Center Jeju Korea.

[42] M. Quigley *et al.*, "ROS: an open-source Robot Operating System," in *ICRA workshop on open source software,* 2009, vol. 3, no. 3.2, p. 5: Kobe, Japan.

[43] G. Falco, M. Pini, and G. Marucco, "Loose and tight GNSS/INS integrations: Comparison of performance assessed in real urban scenarios," *Sensors,* vol. 17, no. 2, p. 255, 2017.

[44] W. Wen, T. Pfeifer, X. Bai, and L. T. Hsu, "Factor Graph Optimization for GNSS/INS Integration: A Comparison with the Extended Kalman Filter (Accepted)," *Navigation,* 2020.

[45] J. Jiao, Y. Yu, Q. Liao, H. Ye, and M. Liu, "Automatic Calibration of Multiple 3D LiDARs in Urban Environments," *arXiv preprint arXiv:1905.04912,* 2019.

[46] C. M. Silva, B. M. Masini, G. Ferrari, and I. Thibault, "A survey on infrastructure-based vehicular networks," *Mobile Information Systems,* vol. 2017, 2017.

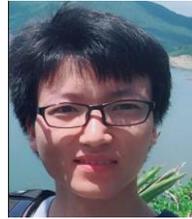**Weisong Wen** was born in Ganzhou, Jiangxi, China. He received a Ph.D. degree in mechanical engineering, the Hong Kong Polytechnic University. He was a visiting student researcher at the University of California, Berkeley (UCB) in 2018. He is currently a research assistant professor in the Department of Aeronautical and Aviation Engineering. His research interests include multi-sensor integrated localization for autonomous vehicles, SLAM, and GNSS positioning in urban canyons.

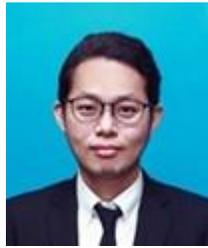**Li-Ta Hsu** received the B.S. and Ph.D. degrees in aeronautics and astronautics from National Cheng Kung University, Taiwan, in 2007 and 2013, respectively. He is currently an assistant professor with the Interdisciplinary Division of Aeronautical and Aviation Engineering, The Hong Kong Polytechnic University, before he served as a post-doctoral researcher in the Institute of Industrial Science at the University of Tokyo, Japan. In 2012, he was a visiting scholar at University College London, the U.K. His research interests include GNSS positioning in challenging environments and localization for pedestrian, autonomous driving vehicle, and unmanned aerial vehicle.